# Class-wise and reduced calibration methods[*]

Michael Panchenko

*appliedAI Institute gGmbH*

m.panchenko@appliedai-institute.de

Anes Benmerzoug

*appliedAI Initiative GmbH*

a.benmerzoug@appliedai.de

Miguel de Benito Delgado[†]

*appliedAI Institute gGmbH*

m.debenito@appliedai-institute.de

*Abstract*—For many applications of probabilistic classifiers it is important that the predicted confidence vectors reflect true probabilities (one says that the classifier is *calibrated*). It has been shown that common models fail to satisfy this property, making reliable methods for measuring and improving calibration important tools. Unfortunately, obtaining these is far from trivial for problems with many classes. We propose two techniques that can be used in tandem. First, a *reduced calibration method* transforms the original problem into a simpler one. We prove for several notions of calibration that solving the reduced problem minimizes the corresponding notion of miscalibration in the full problem, allowing the use of non-parametric recalibration methods that fail in higher dimensions. Second, we propose *class-wise calibration methods*, based on intuition building on a phenomenon called *neural collapse* and the observation that most of the accurate classifiers found in practice can be thought of as a union of $K$ different functions which can be recalibrated separately, one for each class. These typically out-perform their non class-wise counterparts, especially for classifiers trained on imbalanced data sets. Applying the two methods together results in *class-wise reduced calibration algorithms*, which are powerful tools for reducing the prediction and per-class calibration errors. We demonstrate our methods on real and synthetic datasets and release all code as open source in [2, 3].

## I. Introduction

Probabilistic classifiers predict confidence vectors from inputs. Their performance is often evaluated only on the top prediction(s), i.e. on the argmax of the confidences. However, for many decision-making processes, the actual confidence vectors can be relevant. In such cases it is important that confidences are meaningful quantities which, ideally, approximate observed probabilities.

Let $(X, Y)$ be random variables with $X \in \mathcal{X}$, and labels $Y \in \mathcal{Y} := \{1,...,K\}$. For our purposes, a (trained) probabilistic classifier is a deterministic function $c: \mathcal{X} \to \Delta^{K-1}$, where $\Delta^{K-1} := \{x \in [0,1]^K : \sum x_i = 1\}$ is the $K-1$ dimensional simplex. The confidences $C := c(X)$ are a random variable with distribution induced by that of $X$. In what follows we will be mostly concerned with the distribution of $(Y, C)$ and will therefore often omit the dependency on $X$.

Loosely speaking, we call a classifier *calibrated* if at test time the confidence vectors represent true probabilities. More precisely, a classifier $c$ is called *strongly calibrated* iff for every $k \in \{1, \ldots K\}$, $\mathbb{P}(Y = k | C) = C_k$ a.s. For brevity, we write instead[1]

$$\mathbb{P}(Y|C) = C \quad a.s. \tag{1}$$

Unfortunately, as shown in [4], many modern probabilistic classifiers, despite being highly accurate, fail to be strongly calibrated. Given the importance of calibration for practical applications, it is desirable to accurately measure miscalibration as well as to correct for it in some form, e.g. by recalibrating the classifier during or after training ("post-processing").

*Contributions.* Section III proves some elementary but useful bounds for Expected Calibration Error (ECE), which, to our knowledge and despite their simplicity and significance for practitioners, have never been explicitly stated in the literature. Section IV shows how recalibration of $K$-class problems can take place in a much more sample efficient and also computationally cheaper reduced setting while maintaining performance guarantees. Section VI connects calibration and a recently described phenomenon called *neural collapse* [5] which motivates the introduction of a novel algorithm in Section VI that we call *class-wise calibration* and can extend any calibration algorithm. Section VII benchmarks our proposed methods. We release a library of calibration metrics and recalibration methods as open source in [2]. Code to reproduce the experiments can be found in [3].

## II. Previous related work

Calibration has a long history in forecasting, dating at least back to the 1960s [6], and typically revolving around binary predictions in meteorology, using methods like logistic regression to recalibrate probabilities.

In the machine learning community, calibration has traditionally played a lesser role. In the early 2000s, [7] observed that, contrary to margin-based methods, the neural networks of the time predicted calibrated probabilities. For over a decade, calibration received only sporadic attention, until [4] empirically showed that modern features of neural networks (large capacity, batch normalization, etc.) have a detrimental effect on calibration. This work also introduced *temperature scaling*, a uniparametric model for multi-class problems, and showed that it is a maximal entropy solution which efficiently recalibrates in many situations when one is interested in top-class calibration. Immediate extensions to affine maps using vector and matrix scaling can work better when calibration for multiple classes is required but tend to overfit. [8] showed that a generative model of class-conditionals $X|Y$ following Dirichlet distributions is equivalent to Matrix scaling, but provide a probabilistic interpretation.

---

1. For each $k \in \{1, \ldots, K\}$ the left hand side is the *conditional probability* of $Y = k$ given $C$, and it is the r.v. defined as $\mathbb{P}(Y = k|C) := \mathbb{E}[\mathbb{1}_{\{k\}}(Y)|\sigma(C)]$.

[9] empirically show that a variant of the histogram estimator of ECE, which attempts to debias it, and we use in our experiments has better convergence than the standard one. They also introduce a hybrid parametric / non-parametric method with improved error guarantees with respect to scaling and histogram methods.

In the binary setting, [10] use beta distributions and show them to outperform logistic models in many situations, whereas [11] evaluate multiple methods for scoring of loan applications and find non-parametric methods to outperform parametric ones.

Finally, [12] introduce the calibration lens upon which we extend here, highlight the pitfalls of empirical estimates of calibration error and suggest hypothesis testing to overcome some of them.

## III. MEASURING MISCALIBRATION

The *best possible* probabilistic classifier $c^\star$, exactly reproduces the distribution of $Y|X$. With the vector notation introduced above: $\mathbb{P}(Y|X) = c^\star(X)$ a.s. and $c^\star$ is maximally accurate as well as strongly calibrated.[2] For a fixed classifier $c$, and $C := c(X)$, the best one can do is to find a *post-processing* function $r_{\text{id}}: \Delta^{K-1} \to \Delta^{K-1}$ which fulfills[3]

$$r_{\text{id}}(\varsigma) = \mathbb{P}(Y|C = \varsigma) \quad a.s. \quad (2)$$

The composition $r_{\text{id}} \circ c$ is strongly calibrated (although not necessarily accurate) and $r_{\text{id}} = \text{id}$ for any $c$ which is already strongly calibrated. This optimal post-processing function is called *canonical calibration function* and it *gives the best possible post-processing of a probabilistic classifier's outputs*. The goal of any a posteriori recalibration algorithm is to approximate $r_{\text{id}}$.

---

2. To see this, use the tower property: $\mathbb{P}(Y = k|c^\star(X)) = \mathbb{E}[\mathbb{1}_{\{k\}}(Y)|c^\star(X)] = \mathbb{E}[\mathbb{E}[\mathbb{1}_{\{k\}}(Y)|X]|c^\star(X)] = \mathbb{E}[\mathbb{P}(Y = k|X)|\mathbb{P}(Y|X)] = \mathbb{P}(Y = k|X) = c^\star(X)$.

3. Here, $\mathbb{P}(Y|C = \varsigma)$ is a *regular conditional probability* of $Y$ given $C$, which exists e.g. for discrete $Y$ and continuous $C$. The notation $r_{\text{id}}$ is for consistency with the notation introduced in Section IV for calibration lenses.

A natural measure of miscalibration is the expected value of the distance between $r_{\text{id}}$ and id. Given any norm $\|\cdot\|$ over $\Delta^{K-1}$, one defines the *expected strong calibration error* (also *canonical calibration error*) as:

$$\text{ESCE}(c) := \mathbb{E}_C[\|\mathbb{P}(Y|C) - C\|] = \mathbb{E}[\|r_{\text{id}}(C) - C\|].$$

Unfortunately, computing ESCE requires an estimate of $r_{\text{id}}$, and because the latter can be used to recalibrate the classifier, computing ESCE is as hard as recalibrating. Because of this difficulty, practical calibration metrics have to resort to some form of reduction. A common method is to condition on a 1-dimensional projection of $C$, thereby replacing the complicated estimation of a high-dimensional distribution with the much simpler estimation of a 1-dimensional one. The latter can be done e.g. with binning. A general framework for constructing such reductions was introduced by [12] with the concept of calibration lens, see Section IV. Two common examples are *expected (confidence) calibration error*:

$$\text{ECE}(c) := \mathbb{E}_C[|\mathbb{P}(Y = \text{argmax}(C)|\max C) - \max C|], \quad (3)$$

which focuses on the top prediction, and *class-wise ECE*:[4]

$$\text{cwECE}(c) := \frac{1}{K}\sum_{k=1}^{K} \mathbb{E}_C[|\mathbb{P}(Y = k|C_k) - C_k|], \quad (4)$$

which focuses on single classes. For each $k$ we also define $\text{cwECE}_k(c) := \mathbb{E}_C[|\mathbb{P}(Y = k|C_k) - C_k|]$. A strongly calibrated classifier has vanishing ECE and cwECE (as well as all other reductions),[5] but the converse is not true, see [12] for an example. Note that there exist alternative definitions of ECE and cwECE in the literature which condition on $C$ instead of $\max C$ or $C_k$, which are *not the same* as (3) and (4).

In practice we often encounter two important classes of classifiers:

**Definition 1.** *Let* $c: \mathcal{X} \to \Delta^{K-1}$ *be a classifier and* $\tilde{C} := \max C$, $\tilde{Y} := \mathbb{1}_{\{\text{argmax} C\}}(Y)$. *We say that $c$ is* almost always over- *(resp.* under-*) confident if the set* $U := \{\mathbb{P}(\tilde{Y} = 1|\tilde{C}) \leqslant \tilde{C}\}$, *resp.* $U := \{\mathbb{P}(\tilde{Y} = 1|\tilde{C}) \geqslant \tilde{C}\}$, *has* $\mathbb{P}(U) \geqslant 1 - \delta$ *for some* $0 < \delta \ll 1/2$.

Empirically, neural networks are known to be overconfident, making the following bounds of practical significance. They show that to minimize ECE, it is usually enough to achieve high accuracy. Intuitively, an accurate classifier simply does not have much room to be overconfident, and if it is perfectly accurate, it cannot be overconfident at all:

---

4. Following [8], we use a constant factor $1/K$, although it would seem more natural to use weights $1/\mathbb{P}(Y = k)$ instead. Note also that cwECE is an example which is not induced by any calibration lens as introduced in Section IV.

5. To see this, use the tower property as in Footnote 2.

**Lemma 2.** *Let* $\mathrm{acc}_U(c)$ *be the accuracy of $c$ over the set $U$ from Definition 1:*

1. *If $c$ is almost always overconfident, then* $\mathrm{ECE}(c) \leqslant 1 - \mathrm{acc}_U(c)$.

2. *If $c$ is almost always overconfident for a fixed class $k$, then* $\mathrm{cwECE}_k(c) \leqslant 1 - \mathbb{P}(Y = k | U)$.

3. *If $c$ is almost always under-confident for class $k$, then* $\mathrm{ECE}(c) \leqslant \mathrm{acc}_U(c)$, *and* $\mathrm{cwECE}_k(c) \leqslant \mathbb{P}(Y = k | U)$.

*Proof.* *1.* On the set $U$, we can use the linearity of the expectation, and on $U^c$ we can bound the integrand by 1 to obtain: $\mathrm{ECE}(c) \leqslant \mathbb{E}[(\tilde{C} - \mathbb{P}(\tilde{Y} = 1 | \tilde{C})) \mathbb{1}_U] + \delta = \mathbb{E}[\tilde{C} \mathbb{1}_U] - \mathbb{P}(\tilde{Y} = 1 | U) + \delta \leqslant 1 - \delta - \mathrm{acc}_U(c) + \delta$.

*2.* Analogously: $\mathrm{cwECE}_k(c) \leqslant \mathbb{E}[C_k \mathbb{1}_U] - \mathbb{E}[\mathbb{P}(Y = k | C_k) \mathbb{1}_U] + \delta \leqslant 1 - \mathbb{P}(Y = k | U)$.

*3.* Swap the terms in the previous computation. □

## IV. CALIBRATION IN A REDUCED SETTING

As a formalization of the process of focusing on specific aspects of calibration [12] introduce the *calibration lens*. For our purposes, this is a map $\phi \colon \mathcal{Y} \times \Delta^{K-1} \to [m] \times \Delta^{m-1}$, with $[m] := \{0, \ldots, m\}$, generating a reduced problem such that $\phi \colon (y, c) \mapsto (\tilde{y}, \tilde{c})$ with $\tilde{y} = \phi_y(y, c)$ and $\tilde{c} = \phi_c(c)$ fulfills $\tilde{c}_i = \mathbb{P}_{y \sim \mathrm{Cat}(c)}(\tilde{y} = i)$. One of the strongest meaningful reductions one can make is the *confidence lens*:

$$\phi_{\mathrm{conf}}(y, c) := (\mathbb{1}_{\{\mathrm{argmax}\, c\}}(y), \max c) \in \{0, 1\} \times [0, 1],$$

which reduces a $K$-class problem into a binary one.[6] The (reduced) canonical calibration function for this new problem is $r_{\phi_{\mathrm{conf}}}(\varsigma) = \mathbb{P}(\tilde{Y} = 1 | \tilde{C} = \varsigma) = \mathbb{P}(Y = \mathrm{argmax}\, C | \max C = \varsigma)$ and the strong calibration error for the induced problem equals the ECE of the original problem. The induced strong calibration error vanishes when $r_{\phi_{\mathrm{conf}}} = \mathrm{id}$. See Appendix B for more examples of lenses and their properties.

More generally, for any calibration lens $\phi = (\phi_y, \phi_c)$, one has an associated canonical calibration function $r_\phi(\varsigma) := \mathbb{P}(\phi_y | \phi_c = \varsigma)$ and an error based on the distance from $r_\phi$ to the identity:

$$\mathrm{E}\phi\mathrm{E}_p(c) := \mathbb{E}_C[\|\mathbb{P}(\phi_y(Y, C) | \phi_c(C)) - \phi_c(C)\|_p],$$

for any $p$-norm, with $1 \leqslant p \leqslant \infty$ (in the sequel we fix some value of $p$ and omit the subindex). If $\phi = \mathrm{id}$ then $\mathrm{E}\phi\mathrm{E} = \mathrm{ESCE}$.

---

6. Several past works make implicit use of the reduced construction $(\tilde{Y}, \tilde{C})$, e.g. [9] by calibrating one-vs-all models.

The main result of this section shows that if, starting from a recalibration function $\tilde{r}$ for the reduced problem, one can construct another recalibration function $\bar{r}$ fulfilling some mild conditions, then strong calibration error guarantees for the reduced problem translate to error guarantees for the original problem in terms of $\mathrm{E}\phi\mathrm{E}$. Because the reduced problems are designed to be of lower dimension, calibration methods typically perform better, improving calibration for the original problem with higher sample efficiency and reduced computational cost. Especially non-parametric methods which can easily underperform in higher dimensions benefit from reduced calibration[7].

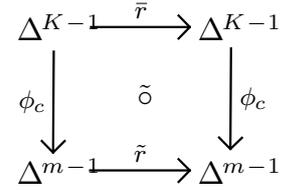

**Figure 1.** Reduced calibration $\bar{r}$ from $\phi_c$ and $\tilde{r}$: if the diagram is commutative with high probability, then ESCE for the reduced problem transfers to $\mathrm{E}\phi\mathrm{E}$ for the original one.

The key intuition about the construction $\tilde{r} \mapsto \bar{r}$ is that it provides a right inverse for $\phi_c$ on most of the predictions, in the sense that $\phi_c \circ \bar{r} = \tilde{r} \circ \phi_c$ holds with high probability, see Figure 1. The extent to which this happens determines how successfully one can lift calibration from a reduced problem to the original one:

**Lemma 3.** *Let $\phi$ be a calibration lens with $\phi(Y, C) \in [m] \times \Delta^{m-1}$. Fix $(\tilde{Y}, \tilde{C}) = \phi(Y, C)$ and assume one has calibration functions $\tilde{r} \colon \Delta^{m-1} \to \Delta^{m-1}$ and $\bar{r} \colon \Delta^{K-1} \to \Delta^{K-1}$ fulfilling: (1) there exists $\delta > 0$ such that the set $U := \{\phi_c \circ \bar{r} = \tilde{r} \circ \phi_c\}$ has $\mathbb{P}(U) \geqslant 1 - \delta$; and: (2) $\tilde{r}$ is "almost calibrated" in the sense that $\|\mathbb{P}(\tilde{Y} | \tilde{r}(\tilde{C})) - \tilde{r}(\tilde{C})\|_U \leqslant \varepsilon$ for some $\varepsilon > 0$. Then:*

$$\mathrm{E}\phi\mathrm{E}(\bar{r} \circ c) \leqslant \varepsilon + \delta.$$

*If, in particular $\mathbb{P}(U) = 1$, then $\mathrm{E}\phi\mathrm{E}(\bar{r} \circ c) = \mathrm{ESCE}(\tilde{r} \circ \tilde{c})$.*

The final observation is of practical relevance for parametric $\tilde{r}$ when it is often possible to compute $U$ exactly. E.g. for temperature scaling $\tilde{r}_T(c) = \sigma(\log(c)/T)$, if $T \geqslant 1$ then $\tilde{r}_T([0, 1]) \subseteq [1/K, 1] \Rightarrow \mathbb{P}(U) = 1$ (see Corollary 4).

**Proof. (of Lemma 3)** Define $Z := \mathbb{P}(\tilde{Y} | \phi_c(\bar{r}(C))) - \phi_c(\bar{r}(C))$. By the construction of $U$, one has $\|Z \mathbb{1}_U\| = \|\mathbb{P}(\tilde{Y} | \tilde{r}(\tilde{C})) - \tilde{r}(\tilde{C})\|_U \leqslant \varepsilon$, and over $U^c$, $\|Z \mathbb{1}_{U^c}\| \leqslant \|Z\| \mathbb{P}(U^c) \leqslant \delta$. Consequently:

$$\begin{aligned}\mathrm{E}\phi\mathrm{E}(\bar{r} \circ c) &= \mathbb{E}[\|\mathbb{P}(\tilde{Y} | \phi_c(\bar{r}(C))) - \phi_c(\bar{r}(C))\|] \\ &= \mathbb{E}[\|Z \mathbb{1}_U + Z \mathbb{1}_{U^c}\|] \leqslant \varepsilon + \delta,\end{aligned}$$

as desired. If $\delta = 0$, then $\mathrm{E}\phi\mathrm{E}(\bar{r} \circ c) = \mathbb{E}[\|Z\|] = \mathbb{E}[\|\mathbb{P}(\tilde{Y} | \tilde{r}(\tilde{C})) - \tilde{C}\|] = \mathrm{ESCE}(\tilde{r} \circ \tilde{c})$. □

---

7. A notable exception where reducing the problem may go wrong is temperature scaling. As we will show experimentally in Section VII, for certain 1-dim. calibration problems temperature scaling fails to give a good approximation.

The following corollary provides our first practical result with an explicit recalibration function constructed from a reduced binary calibration method:

**Corollary 4. (Reduced confidence calibration)** *Let $\phi_{\text{conf}}(y,c) = (\mathbb{1}_{\{\operatorname{argmax} c\}}(y), \max c)$ and $\tilde{r}: [0,1] \to [0,1]$ with $\tilde{U} := \{\tilde{r}(\max C) \geqslant 1/K\}$. Define*

$$\bar{r}(c) := \tilde{r}(c_a)\,\vec{e}_a + \sum_{i \neq a} \frac{1 - \tilde{r}(c_a)}{K-1}\,\vec{e}_i,$$

*where $a = \operatorname{argmax} c$ and $c \in \Delta^{K-1}$. If $\mathbb{P}(\tilde{U}) \geqslant 1 - \delta$ and $\|\mathbb{P}(\tilde{Y}|\tilde{r}(\tilde{C})) - \tilde{r}(\tilde{C})\|_U \leqslant \varepsilon$, then*

$$\operatorname{ECE}(\bar{r} \circ c) \leqslant \varepsilon + \delta.$$

**Proof.** On $\tilde{U}$ we have $\tilde{r}(\max C) \geqslant \frac{1}{K}$ and then $\frac{1 - \tilde{r}(\max C)}{K-1} \leqslant \frac{1}{K}$. Consequently, $\max \bar{r}(C) = \tilde{r}(\max C)$, and $\operatorname{argmax} \bar{r}(C) = \operatorname{argmax} C$ over $\tilde{U} \subseteq U$ (in fact, here $\tilde{U} = U$ holds), and we can apply Lemma 3 to conclude. □

*About the condition $\tilde{r}(\max C) \geqslant 1/K$ in Corollary 4*: If $c$ is such that $\mathbb{P}(Y = \operatorname{argmax} C | \max C) < 1/K$, then this is clearly a "poor" classifier. One would at least expect that the predicted top class of $c(X)$ agrees with the class $k$ such that $\mathbb{P}(Y = k|C)$ is highest. If $r_{\text{id}}$ is almost argmax-preserving (see (6)), there exists $\delta > 0$ such that the set $A := \{\operatorname{argmax} r_{\text{id}}(C) = \operatorname{argmax} C\}$, has $\mathbb{P}(A) \geqslant 1 - \delta$. In this set one has $\mathbb{P}(Y = \operatorname{argmax} C|C) = \mathbb{P}(Y = \operatorname{argmax} r_{\text{id}}(C)|C) = \max r_{\text{id}}(C)$ with the last equality following from the fact that $r_{\text{id}}$ is the canonical calibration function. Using this and the tower property we can compute, always in the set $A$,

$$\begin{aligned}
\mathbb{P}(\tilde{Y} = 1 | \tilde{r}(\tilde{C})) &= \mathbb{E}[\mathbb{1}_{\{\operatorname{argmax} C\}}(Y) | \tilde{r}(\tilde{C})] \\
&= \mathbb{E}[\mathbb{E}[\mathbb{1}_{\{\operatorname{argmax} C\}}(Y) | C] | \tilde{r}(\tilde{C})] \\
&= \mathbb{E}[\mathbb{P}(Y = \operatorname{argmax} C|C) | \tilde{r}(\tilde{C})] \\
&= \mathbb{E}[\max r_{\text{id}}(C) | \tilde{r}(\tilde{C})] \\
&\geqslant \frac{1}{K}.
\end{aligned}$$

If now we have an over-confident recalibrated classifier $\tilde{r} \circ \tilde{C}$, $\tilde{r}(\tilde{C}) \geqslant \mathbb{P}(\tilde{Y} = 1 | \tilde{r}(\tilde{C})) \geqslant 1/K$. The natural property of a "good" classifier $C$ to induce a canonical calibration function that is argmax-preserving, together with an assumption on the confidence of $\tilde{r} \circ \tilde{C}$ yield the condition required in Lemma 3.

## V. NEURAL COLLAPSE AND CALIBRATION

One of the results of recent work on *neural collapse* [13, 5] is that the activations returned by the penultimate layer of a deep neural classifier as well as the weights of the last, classifying layer align themselves with an *Equiangular Tight Frame* (ETF). Apart from the rotation of the frame, which does not influence the predictions, and the center, which is compensated by the classifier's bias vector, the ETF has only one free parameter - its radius. We show that it is tightly related to calibration and temperature scaling. The observed convergence to the ETF will serve as a motivation for defining class-wise calibration methods below.

From [5, Section 5A] follows that a sample $x$ with label $i$ will be approximately mapped to the logits vector $l(x)$

$$l(x) \approx R_{\text{act}} R_{\text{wts}} \left( \vec{e}_i - \frac{1}{K-1} \sum_{j \neq i} \vec{e}_j \right) + \frac{1}{K} \sum_j \vec{e}_j, \quad (5)$$

where $R_{\text{wts}}$ is the norm of the classifier's weights (which form an ETF) and $R_{\text{act}}$ is the radius of the ETF formed by the activations of the penultimate layer (for a detailed derivation see Appendix A). Thus the radii of the ETFs directly control the confidences predicted for training data.[8] For a negative log-likelihood (n.l.l.) loss, optimisation increases $R_{\text{act}} R_{\text{wts}}$ towards infinity, even after convergence to the ETFs has occurred. This leads to an overfitting of $R_{\text{act}}$ and $R_{\text{wts}}$ on the training set, explaining why on the test set neural networks are generally overconfident and why regularizers like weight decay combat this by preventing $R_{\text{act}} R_{\text{wts}}$ from growing unboundedly.

Since temperature scaling multiplies the logits vector with a single parameter, it is equivalent to adjusting $R_{\text{wts}}$, overcoming overfit. *It is also the only recalibration method that leaves the ETF structure of predictions intact by construction*. E.g. vector scaling would violate the equinorm property of an ETF if all the parameters scaling the logits are not equal. The fact that convergence of activations and classifier-layer rows to an ETF is beneficial (see [5]), provides an explanation for why vector scaling trained with n.l.l. tends to converge to temperature scaling. And also for the success of temperature scaling: in order to keep the ETF structure intact, the only free parameter influencing the predictions that one can recalibrate is $R_{\text{wts}}$, which is equivalent to temperature scaling.

Many parametric recalibration methods, like vector scaling, beta calibration or Dirichlet scaling find a transformation of the logits by optimizing a variant of n.l.l. on a dataset that contains *representatives for all classes*. In such a situation, assuming that neural collapse is favored by n.l.l. and keeping in mind that recalibration methods do not adjust the outputs of the penultimate layer, one must conclude that these algorithms have an inductive bias towards an equal treatment of all classes. This inductive bias is strongly related to the one described in [14].

However, the amount of overconfidence due to overfitting of the ETF radii may differ from class to class and it would be useful to have a recalibration method that is able to accommodate such differences, i.e. without the inductive bias of converging to an ETF. Below we show that a simple modification of recalibration algorithms, which we call *class-wise calibration*, addresses this problem. In class-wise calibration only representatives of one class at a time form part of the optimization problem and thus the inductive bias towards convergence to an ETF is avoided. We verify experimentally that class-wise calibration is generally beneficial for parametric recalibration methods like temperature scaling and beta calibration in Section VII. It also helps with non-parametric methods like histogram binning and isotonic regression, although to a lesser extent.

---

[8]. The term $\frac{1}{K}\mathbb{1}_K$ is cancelled by the softmax and thus irrelevant. [5] use the factor $\alpha = \frac{R_{\text{wts}}}{R_{\text{act}}}$ in calculations of the argmax of the logits.

## VI. CLASS-WISE CALIBRATION

As stated above, the goal of any calibration algorithm is to approximate the canonical calibration function $r_{\text{id}}$ (or its reduction) with some $r$. However, even for strong calibration, the differences between $r_{\text{id}}$ and $r$ are not equally important across $\Delta^{K-1}$. Since all calibration errors relevant for practical purposes rely on expected values, it is only important that $r$ *be a good approximation to $r_{\text{id}}$ in regions of high probability w.r.t. the distribution of $C$*.

It is empirically observed that accurate classifiers, especially overconfident ones like neural networks, tend to output confidence vectors that are close to the corners of $\Delta^{K-1}$. Therefore, and to prevent an inductive bias towards an ETF, it seems reasonable to split $\Delta^{K-1}$ into $K$ sectors

$$S_k := \{c \in \Delta^{K-1} | \text{argmax}(c) = k\} \text{ for } k \in \{1, \ldots, K\},$$

and find independent approximations of $r_{\text{id}}$ on each $S_k$ separately instead of trying to find a single approximation that works well on all of them simultaneously. The basic idea is then:

**Algorithm (Class-wise calibration)**

1. Choose $K$ calibration estimators, $\rho_1, \ldots, \rho_K$.
2. Fit $\rho_k$ to $(Y, C) | \text{argmax}(C) = k$.
3. The approximation to $r_{\text{id}}$ is then given by

$$r_{\text{cw}}(c) := \rho_{\text{argmax}(c)}(c).$$

The approximation so obtained will usually present discontinuities at the boundaries of the $S_k$, where it might thus fail to approximate the true $r_{\text{id}}$. However, by assumption these regions have low probability and do not affect expected calibration errors.

Being a strict generalization, class-wise calibration offers more representational power and flexibility than normal calibration, because the different sectors might have entirely different statistical properties (the mistakes in a model's confidences for some class are not necessarily correlated to the mistakes of confidences in another class).

Another heuristic argument to use class-wise calibration is as follows. We expect reasonably good probabilistic classifiers to have the property that the predicted value of $Y$ from an input $X$ is indeed the most probable value of $Y|X$ (even if the classifier is not confidence calibrated). This statement is equivalent to $\text{argmax}(r_{\text{id}}(c(X))) = \text{argmax}(c(X))$, and "sufficiently good" classifiers will satisfy this equation with high probability. I.e. we expect that there is some small $\delta > 0$ s.t.

$$\mathbb{P}(\text{argmax}(r_{\text{id}}(C)) = \text{argmax}(C)) \geq 1 - \delta, \quad (6)$$

and then we say that $r_{\text{id}}$ is *almost argmax-preserving*. If this holds, then $r_{\text{id}}(S_k) \subset S_k$ with high probability, and therefore it makes sense to approximate $r_{\text{id}}$ on each sector separately by an *argmax-preserving $r_k$*.

Any argmax-preserving $r$ leaves the predicted class unaltered and thus $\text{acc}(C) = \text{acc}(r \circ C)$. If each estimator $r_k$ is argmax-preserving, as would be the case for e.g. temperature scaling with $T \geq 1$, so is $r_{\text{cw}}$. This is a convenient property since one does not need to worry about worsening accuracy as can easily happen with high-capacity calibration algorithms due to overfitting. More expressive extensions of temperature scaling that fit more than one parameter, like vector scaling or matrix / Dirichlet scaling [8], are not argmax-preserving and can reduce accuracy. However, class-wise temperature scaling offers higher expressivity than temperature scaling ($K$ parameters instead of one) while not altering the predicted class.

Class-wise calibration can be used together with reduced calibration, thereby alleviating the latter's ignorance about the existence of multiple classes. In the following we demonstrate experimentally that class-wise calibration improves calibration according to various metrics, confidence-reduced calibration tends to improve ECE and the combination of both can further improve calibration across different metrics.

## VII. EXPERIMENTS

We compare multiple calibration algorithms with their reduced, class-wise and class-wise reduced counterparts. Class-wise calibration proves especially useful for imbalanced data sets. The reduced calibration algorithm constructed from a selected calibration lens allows to focus on improving a selected calibration error, e.g. the ECE, instead of trying to improve strong calibration. Moreover, it allows using methods that work well in lower dimensions, like non-parametric methods, for higher dimensional problems.

A major hurdle for the comparison is the poor sample efficiency of the estimators of expected calibration metrics. To understand the problem consider a typical RESNET16 [15] trained on the CIFAR-10 dataset [16], where the test set contains 10k samples. With 5-fold cross-validation each fold has 2000 samples to estimate a metric. For cwECE, there will be only around 200 of them per class. Since the distribution of confidences predicted by RESNET16 is very skewed towards the vertices of the simplex, most bins will contain very few samples (in an equal-size binning scheme), leading to high variance of the cwECE estimator and poorly interpretable results. A similar (albeit not so dramatic) effect can be observed for ECE, as we show in the figures and tables with real data below.

Therefore, to test with enough samples, we first work with a random forest trained on two synthetic datasets with 5 classes and 60k samples, where one of the datasets is imbalanced.[9] The model is trained on 30k samples (from a stratified shuffle split) and reaches an accuracy of roughly 89% in both cases. As is common with random forests, the resulting model is highly miscalibrated (ECE $\approx 0.23$).

---

9. We used SCIKIT-LEARN's `make_classification(n_samples = 60000, n_classes = 5, n_informative = 15, random_state = 16)` for the balanced data set and the flag `weights=(0.3, 0.1, 0.25, 0.15)` for the imbalanced one.

We evaluate calibration algorithms by performing a 6-fold cross validation on the 30k test samples. Each baseline algorithm is compared against its reduced, class-wise and class-wise reduced versions. Note that training binary recalibration algorithms in a 1-vs.-all fashion (as is done in the baselines for logistic calibration and beta calibration) is not the same as training their reduced version proposed above. The histogram-binning recalibration always used 20 bins.

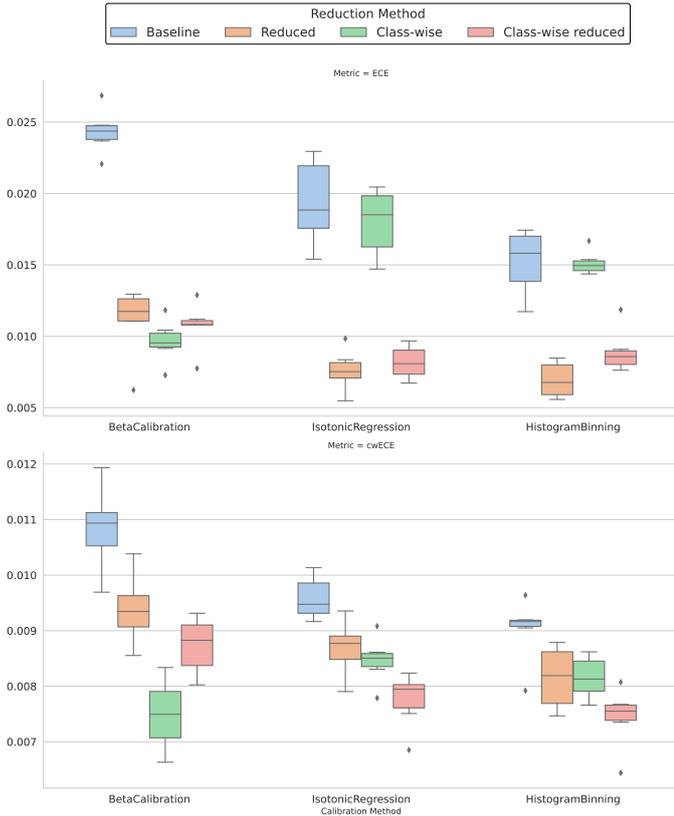

**Figure 2.** Comparison of beta calibration, isotonic regression and histogram binning on ECE / cwECE for a random forest model trained on a balanced synthetic dataset (6 folds; 25 bins).

We see in Figure 2 and Figure 3 that in all cases at least one of the proposed methods outperforms the baseline. The performance boost is particularly striking for non-parametric methods. Because temperature scaling fails at solving the reduced problem well we plot its evaluation separately in the supplementary material.

For a balanced dataset, class-wise calibration provides only a marginal improvement. The situation changes for imbalanced data, where reduced calibration still decreases ECE significantly wrt. the baseline but performs poorly wrt. cwECE, being significantly worse than the baseline. This is natural since the reduced problem loses information about the different classes. However, the class-wise reduced algorithms do a good job in decreasing both the ECE and the cwECE, thereby offering a balance between calibration of confidences in predictions as well as in non-predicted classes.

We repeat the analysis using four models trained on four real-world datasets:

- DEIT [17] pre-trained on IIT-CDIP [18] and then fine-tuned on RVL-CDIP [19], a subset of the former consisting of 400000 grayscale document images split evenly across 16 classes.

- DISTILBERT [20] trained by distilling the BERT base model and then fine-tuning on the IMDB dataset [21], a dataset for binary sentiment classification consisting of 50000 movie reviews.

- LIGHTGBM [22] trained on the SOREL-20M dataset [23], a binary classification dataset consisting of nearly 20 million malicious and benign portable executable files with pre-extracted features and metadata, and high-quality labels.

- RESNET56 [15] trained on CIFAR10 [16], a multi-class classification dataset consisting of 60000 images split evenly across 10 classes.

Due to the relatively small number of samples in the test set of the CIFAR10 dataset we only use 4-fold cross validation and 20 bins to estimate ECE and cwECE. Our results are summarized in Table-1 and Table-2. Note how in almost every case one of our proposed methods outperforms the baseline.

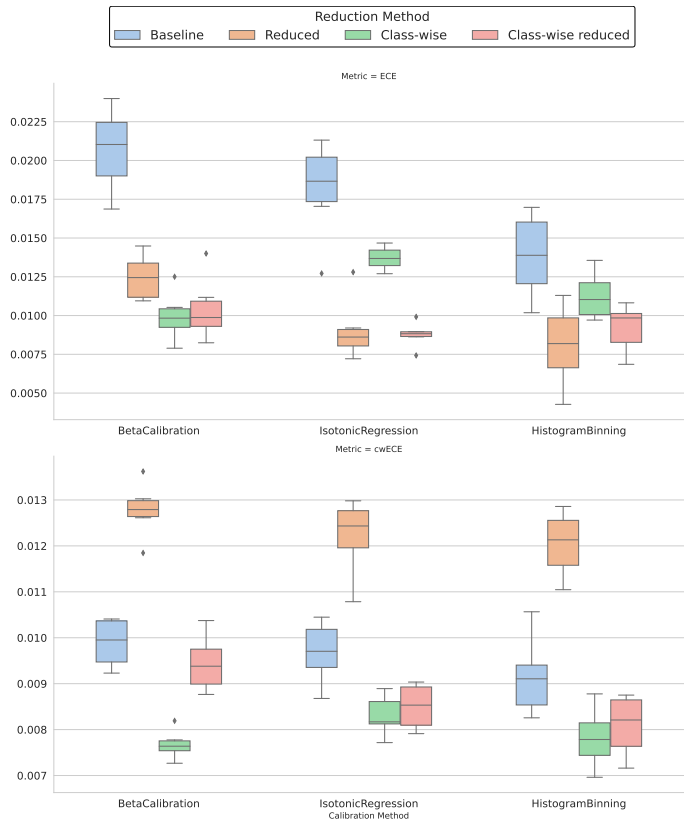

**Figure 3.** Comparison of beta calibration, isotonic regression and histogram binning on ECE / cwECE for a random forest model trained on an imbalanced synthetic dataset (6 folds; 25 bins).

| Model | Method | Baseline | Reduced | Class-wise | Class-wise reduced |
|---|---|---|---|---|---|
| DEIT | TempScaling | 0.02902 ± 6% | -11.85% ± 5% | -27.17% ± 10% | **-37.41%** ± 10% |
| | Beta | 0.02504 ± 11% | -23.07% ± 6% | -36.07% ± 12% | **-38.36%** ± 9% |
| | Isotonic | 0.00775 ± 24% | -23.90% ± 13% | +6.64% ± 22% | -8.11% ± 15% |
| | Histogram | 0.01782 ± 9% | **-72.09%** ± 5% | -9.92% ± 11% | -56.96% ± 8% |
| DISTILBERT | TempScaling | 0.01921 ± 22% | 0.00% ± 22% | -5.21% ± 30% | -5.21% ± 30% |
| | Beta | 0.01508 ± 27% | -10.78% ± 18% | +30.45% ± 53% | +30.45% ± 53% |
| | Isotonic | 0.06002 ± 29% | -82.08% ± 4% | +1.53% ± 28% | +1.53% ± 28% |
| | Histogram | 0.11305 ± 11% | **-88.28%** ± 3% | 0.00% ± 11% | 0.00% ± 11% |
| LIGHTGBM | TempScaling | 0.00317 ± 4% | 0.00% ± 4% | -15.33% ± 4% | -15.33% ± 4% |
| | Beta | 0.00223 ± 4% | -8.46% ± 6% | -6.03% ± 4% | -6.05% ± 4% |
| | Isotonic | **0.00030** ± 52% | +13.23% ± 32% | +0.30% ± 52% | +0.30% ± 52% |
| | Histogram | **0.00026** ± 31% | +22.73% ± 31% | 0.00% ± 31% | 0.00% ± 31% |
| RESNET56 | TempScaling | 0.01159 ± 8% | +2.22% ± 29% | -12.86% ± 18% | +21.80% ± 17% |
| | Beta | **0.01198** ± 20% | -7.14% ± 19% | +15.42% ± 23% | +7.19% ± 18% |
| | Isotonic | 0.01270 ± 12% | -13.70% ± 22% | +38.24% ± 15% | +12.96% ± 23% |
| | Histogram | 0.01843 ± 19% | **-67.76%** ± 7% | -16.38% ± 15% | -24.83% ± 16% |
| RF BALANCED | TempScaling | **0.02636** ± 8% | +416.88% ± 13% | +0.70% ± 8% | +419.02% ± 14% |
| | Beta | 0.02435 ± 6% | -54.54% ± 10% | -60.46% ± 6% | -56.03% ± 7% |
| | Isotonic | 0.01936 ± 16% | -60.67% ± 8% | -7.04% ± 12% | -57.81% ± 6% |
| | Histogram | 0.01523 ± 15% | **-54.45%** ± 8% | -0.60% ± 5% | -41.35% ± 10% |
| RF IMBALANCED | TempScaling | 0.02469 ± 8% | +364.35% ± 16% | -2.89% ± 7% | +367.57% ± 19% |
| | Beta | 0.02070 ± 13% | -39.78% ± 7% | -51.90% ± 7% | -49.77% ± 10% |
| | Isotonic | 0.01817 ± 17% | -50.13% ± 11% | -24.58% ± 4% | **-51.75%** ± 4% |
| | Histogram | 0.01385 ± 20% | **-41.69%** ± 19% | -18.72% ± 11% | -33.42% ± 11% |

**Table 1.** Summary of ECE scores for DEIT, DISTILBERT, LIGHTGBM, RESNET56 and Random Forest. All values are relative to the baseline with ±1 stdev. The lowest average ECE score per calibration method is highlighted in **bold** and the corresponding cell in grey. The best ECE score per model is highlighted as white text over a black background.

| Model | Method | Baseline | Reduced | Class-wise | Class-wise reduced |
|---|---|---|---|---|---|
| DEIT | TempScaling | 0.00394 ± 6% | +5.15% ± 5% | **-9.64%** ± 3% | -3.50% ± 4% |
| | Beta | 0.00319 ± 5% | +25.72% ± 5% | **-11.55%** ± 7% | +17.49% ± 3% |
| | Isotonic | **0.00266** ± 5% | +29.72% ± 4% | +0.67% ± 4% | +15.40% ± 6% |
| | Histogram | 0.00291 ± 4% | +15.86% ± 4% | **-5.13%** ± 3% | +7.25% ± 5% |
| DISTILBERT | TempScaling | **0.11227** ± 13% | 0.00% ± 13% | +15.07% ± 6% | +15.07% ± 6% |
| | Beta | 0.13535 ± 3% | -16.46% ± 12% | +0.61% ± 3% | +0.61% ± 3% |
| | Isotonic | 0.13689 ± 3% | -16.82% ± 12% | +0.02% ± 3% | +0.02% ± 3% |
| | Histogram | 0.14230 ± 2% | -17.49% ± 9% | 0.00% ± 2% | 0.00% ± 2% |
| LIGHTGBM | TempScaling | 0.00583 ± 2% | 0.00% ± 2% | -33.21% ± 2% | -33.21% ± 2% |
| | Beta | 0.00267 ± 5% | +106.36% ± 8% | -6.31% ± 4% | -6.29% ± 4% |
| | Isotonic | **0.00038** ± 35% | +1290.63% ± 53% | +0.48% ± 35% | +0.48% ± 35% |
| | Histogram | **0.00037** ± 28% | +1368.65% ± 52% | 0.00% ± 28% | 0.00% ± 28% |
| RESNET56 | TempScaling | 0.00746 ± 2% | -5.80% ± 7% | -4.02% ± 5% | -5.90% ± 6% |
| | Beta | **0.00687** ± 2% | +7.98% ± 8% | +4.25% ± 5% | +8.32% ± 6% |
| | Isotonic | 0.00671 ± 3% | +4.86% ± 9% | +10.35% ± 10% | -3.32% ± 7% |
| | Histogram | 0.00682 ± 6% | +0.58% ± 7% | **-16.71%** ± 5% | -6.88% ± 5% |
| RF BALANCED | TempScaling | **0.01228** ± 7% | +339.01% ± 10% | +0.35% ± 6% | +336.59% ± 10% |
| | Beta | 0.01085 ± 7% | -13.45% ± 6% | **-30.96%** ± 6% | -19.47% ± 5% |
| | Isotonic | 0.00958 ± 4% | -9.32% ± 5% | -11.66% ± 4% | **-19.06%** ± 5% |
| | Histogram | 0.00902 ± 6% | -9.63% ± 6% | -9.60% ± 4% | **-17.54%** ± 6% |
| RF IMBALANCED | TempScaling | 0.01746 ± 2% | +160.88% ± 1% | -9.15% ± 4% | +165.54% ± 10% |
| | Beta | 0.00989 ± 5% | +29.19% ± 6% | **-22.44%** ± 3% | -4.59% ± 6% |
| | Isotonic | 0.00969 ± 7% | +26.15% ± 8% | -14.28% ± 5% | -12.20% ± 5% |
| | Histogram | 0.00915 ± 9% | +31.66% ± 8% | -14.55% ± 7% | -11.51% ± 7% |

**Table 2.** Summary of cwECE score for DEIT, DISTILBERT, LIGHTGBM, RESNET56 and Random Forest. All values are relative to the baseline with ±1 stdev. The lowest average cwECE score per calibration method is highlighted in **bold** and the corresponding cell in grey. The best cwECE score per model is highlighted as white text over a black background.

## APPENDIX A. DERIVATION OF EQUATION 5

From the analysis in [5, Section 5A] we have that a fully trained neural net maps an activation vector $h$ of the penultimate layer to approximately

$$\alpha\, M^\top (h - \mu_G) + \frac{1}{K} \sum_i \vec{e}_i,$$

where $M^\top$ is a matrix formed by stacking the rows of means of activations $\mu_k - \mu_G, k = 1, \ldots, K,$ and $\alpha$ fulfills

$$W = \alpha\, M^T \Longrightarrow \alpha = \frac{R_{\text{wts}}}{R_{\text{act}}},$$

where $W$ are the weights of the classifier, $R_{\text{wts}}$ is the euclidian norm of one row of $W$, and $R_{\text{act}}$ the norm of one row of $M^T$ (all rows have the same length due to the ETF property). Since $h$ is the vector of activations computed from a sample that was drawn from the same distribution as the training data, the last linear layer will map to a vector close to $\mu_i$, with $i$ being the corresponding label. Using

$$\langle \mu_k - \mu_G, \mu_i - \mu_G \rangle = R_{\text{act}}^2 \left( \frac{K}{K-1} \delta_{ki} - \frac{1}{K-1} \right),$$

we arrive at (5).

## APPENDIX B. OTHER LENSES

**Notation 5.** *In this section, let $\text{ord}(c) = (c_{i_1}, \ldots, c_{i_K})$ be the ordinal statistic of $c$ and define $\text{arank}(c) = (i_1, \ldots, i_K)$ where the $i_j$ are defined by $\text{ord}(c)$. We denote with $v_{:k}$ the first $k$ components of any vector $v$.*

Besides the confidence lens introduced in Section IV, another possible reduction is that of *top-$k$ calibration*. The lens

$$\phi_k(y, c) := \Big( \sum_{i=1}^k i\, \mathbb{1}_{\{\text{arank}(c)_i\}}(y), \text{ord}(c)_i \Big) \in [k] \times [0,1]^{k-1},$$

has the associated *top-$k$ confidence error*

$$\text{ECE}_k(c) := \mathbb{E}[\|\mathbb{P}(Y = \text{arank}_{:k}(C)|\text{ord}_{:k}(C)) - \text{ord}_{:k}(C)\|].$$

**Corollary 6. (Reduced top-$k$ calibration)** *Let $\phi_k(y, c) := (\sum_{i=1}^k \mathbb{1}_{\{\text{arank}_i(c)\}}(y), \text{ord}_{:k}(c)) = (\tilde{y}, \tilde{c})$, $\tilde{r} \colon \Delta^{k-1} \to \Delta^{k-1}$, $\tilde{U} := \bigcap_{j=1}^k \{\tilde{r}_j(\tilde{C}) \geq 1/K\}$, and*

$$\bar{r}(c) := \sum_{j=1}^k \tilde{r}_j(\tilde{c})\, \vec{e}_{i_j} + \sum_{i \notin \text{arank}(c)} \frac{1 - \sum \tilde{r}_j(\tilde{c})}{K - k} \vec{e}_i,$$

*where $i_j = \text{arank}_j(c)$ and $c \in \Delta^{K-1}$. If $\mathbb{P}(U) \geq 1 - \delta$ and $\|\mathbb{P}(\tilde{Y}|\tilde{r}(\tilde{C})) - \tilde{r}(\tilde{C})\|_U \leq \varepsilon$, then $\mathrm{E}\phi\mathrm{E}(\bar{r} \circ c) \leq \varepsilon + \delta$.*

**Proof.** On $\tilde{U}$ we have $\frac{1 - \sum \tilde{r}_j(\tilde{C})}{K - k} \leq \frac{1 - k/K}{K - k} \leq \frac{1}{K}$, hence $\text{ord}_{:k}(\bar{r}(C)) = \tilde{r}(\text{ord}_{:k}(C))$ and $\text{arank}(\bar{r}(C)) = \text{arank}(C)$, and $\tilde{U} \subseteq U$. □

Yet another interesting reduction is provided by the total confidence in the top $k$ classes:

**Corollary 7. (Reduced sum-$k$ calibration)** *Let $\phi_{\Sigma k}(y, c) := (\mathbb{1}_{\{\text{arank}(c)\}}(y), \sum_{i=1}^k \text{ord}_i(c)) = (\tilde{y}, \tilde{c})$, $\tilde{r} \colon [0, 1] \to [0, 1]$, $\tilde{U} := \{\tilde{r}(\tilde{C}) \geq k/K\}$, and*

$$\bar{r}(c) := \sum_{j=1}^k \frac{\tilde{r}(\tilde{c})}{k} \vec{e}_{i_j} + \sum_{i \notin \text{arank}(c)} \frac{1 - \tilde{r}(\tilde{c})}{K - k} \vec{e}_i$$

*where $i_j = \text{arank}_j(c)$ and $c \in \Delta^{K-1}$. If $\mathbb{P}(U) \geq 1 - \delta$ and $\|\mathbb{P}(\tilde{Y}|\tilde{r}(\tilde{C})) - \tilde{r}(\tilde{C})\|_U \leq \varepsilon$, then $\mathrm{E}\phi\mathrm{E}(\bar{r} \circ c) \leq \varepsilon + \delta$.*

**Proof.** On $\tilde{U}$ we have $\frac{1 - \tilde{r}(\tilde{C})}{K - k} \leq \frac{1}{K} \leq \tilde{r}(\tilde{C})$, hence $\text{arank}\,\bar{r}(C) = \text{arank}\, C$ and $\sum_{j=1}^k \text{ord}_j(\bar{r}(C)) = \tilde{r}(\tilde{C})$ and $\tilde{U} \subseteq U$. □

## APPENDIX C. REDUCED TOP-LABEL CALIBRATION

In independent work recently published, [24] reason that classifiers not only predict confidences but also classes and that being confidence-calibrated is hard to interpret and uninformative [24, Example 1]. We can express their construction in the framework of calibration lenses by modifying our simplified definition of lens to be a map $\hat{\phi} \colon \mathcal{Y} \times \Delta^{K-1} \to \{0,1\} \times [0,1] \times [K]$, such that $\hat{\phi} \colon (y, c) \mapsto (\tilde{y}, \tilde{c}, \tilde{a}) = (\mathbb{1}_{\{\arg\max c\}}(y), \max c, \arg\max c)$. The canonical calibration function for this lens is $r_{\hat{\phi}}(\varsigma, \alpha) = \mathbb{P}(Y = \arg\max C \mid \max C = \varsigma, \arg\max C = \alpha)$. We thus recover the definitions in [24] of *top-label calibration* $P(Y = \arg\max C \mid \max C, \arg\max C) = \max C$, and *top-label-ECE*: $\text{TL-ECE} := \mathbb{E}[|P(Y = \arg\max C \mid \max C, \arg\max C) - \max C|] = E\hat{\phi}E_p(c)$. Then Lemma 3, with $\tilde{r} \colon \Delta^{m-1} \times [K] \to \Delta^{m-1} \times [K]$, and Corollary 4 apply to their setting.

## APPENDIX D. ADDITIONAL PLOTS

Here we plot the remaining results summarised in Table-1 and Table-2 of the main paper.

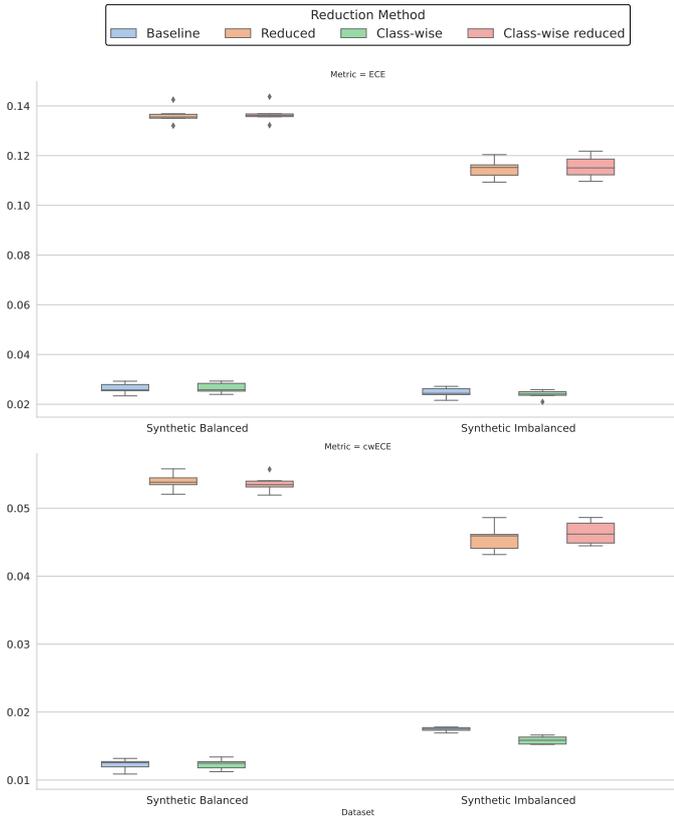

**Figure 4.** Evaluation of Temperature Scaling on ECE / cwECE for a random forest model trained on synthetic datasets (6 folds; 25 bins). For this specific problem, the distribution of the confidences hinders the method from correctly recalibrating, independently of the reduction.

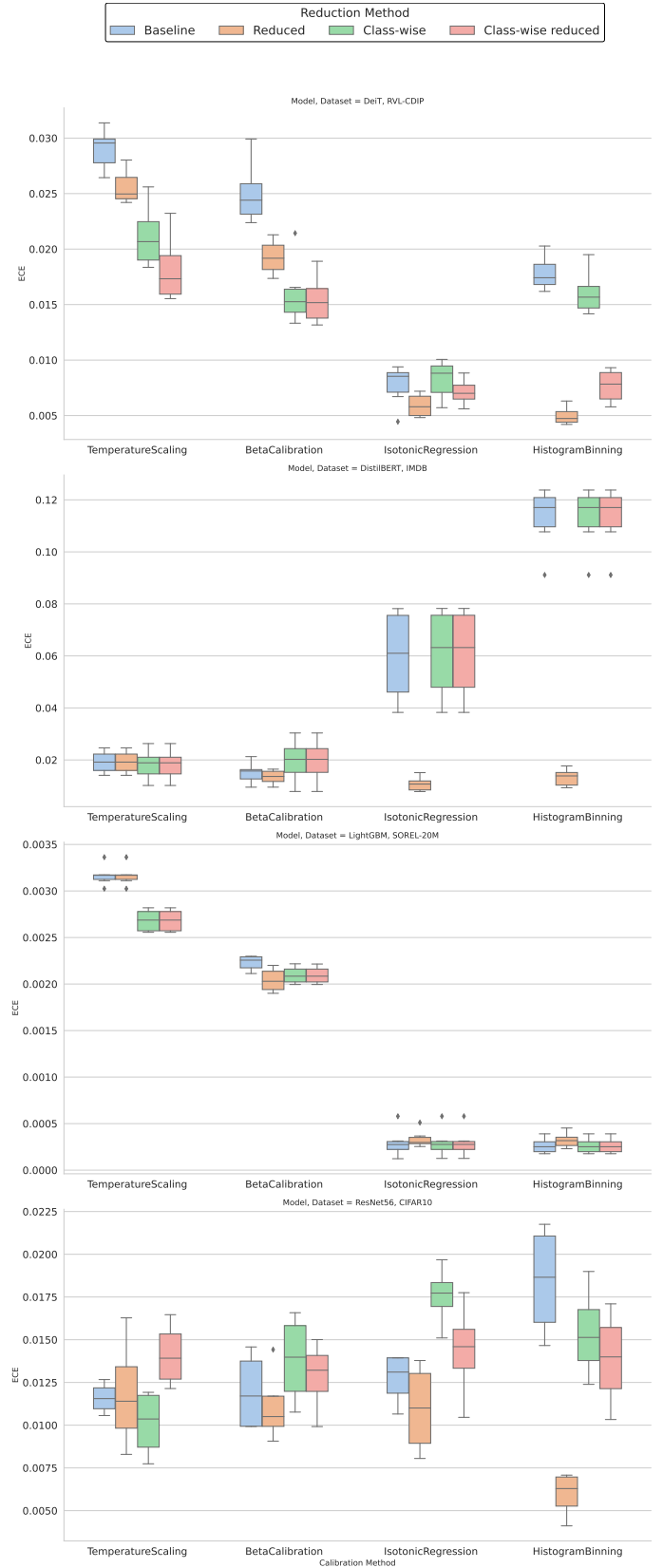

**Figure 5.** Comparison of beta calibration, isotonic regression, histogram binning and temperature scaling on ECE for 4 models trained on 4 different real world datasets (6 folds; 25 bins).

## APPENDIX E. REDUCED AND WEIGHTED REDUCED METHODS

We can improve the strong calibration of Corollary 4 while largely preserving its confidence calibration by weighting the components of $\bar{r}$ with information from the classifier (typically extracted as statistics from a test set):

**Corollary 8.** (Reduced weighted confidence calibration) Let $\phi_{\text{conf}}(y,c) = (\mathbb{1}_{\{\arg\max c\}}(y), \max c)$, $\tilde{r}: [0,1] \to [0,1]$, and $\tilde{U} := \left\{ \tilde{r}(\max C) \geqslant \frac{C_i}{C_i + \sum_{j \neq a(C)} C_j}, \forall i \neq a(C) \right\}$, where $a(c) := \arg\max_j \{c_j\}$. Let

$$\bar{r}_{\text{weighted}}(c) := \tilde{r}(c_a)\,\vec{e}_a + \sum_{i \neq a} \frac{c_i}{\sum_{j \neq a} c_j} (1 - \tilde{r}(c_a))\,\vec{e}_i,$$

for $a = a(c)$ and $c \in \Delta^{K-1}$. If $\mathbb{P}(\tilde{U}) \geqslant 1 - \delta$ and $\|\mathbb{P}(\tilde{Y}|\tilde{r}(\tilde{C})) - \tilde{r}(\tilde{C})\|_U \leqslant \varepsilon$, then

$$\text{ECE}(\bar{r}_{\text{weighted}} \circ c) \leqslant \varepsilon + \delta.$$

**Proof.** On $\tilde{U}$ we have $\tilde{r}(\max C) \geqslant \frac{C_i}{C_i + \sum_{j \neq a(C)} C_j}$. Rearranging we arrive at: $C_i \frac{1-\tilde{r}(\max C)}{\sum_{j \neq a(C)} C_j} \leqslant \tilde{r}(\max C)$ and, consequently, $\max \bar{r}(C) = \tilde{r}(\max C)$, hence $\arg\max \bar{r}_{\text{weighted}}(C) = \arg\max C$ over $\tilde{U} \subseteq U$, and we can apply Lemma 3 to conclude. □

Note that the $\delta$ in the construction of $\tilde{U}$ for $\bar{r}_{\text{weighted}}$ is equal or larger to the $\delta$ for the non-weighted $\bar{r}$. Thus, while calibration w.r.t. other metrics may improve in the weighted version, calibration w.r.t. ECE can only worsen.

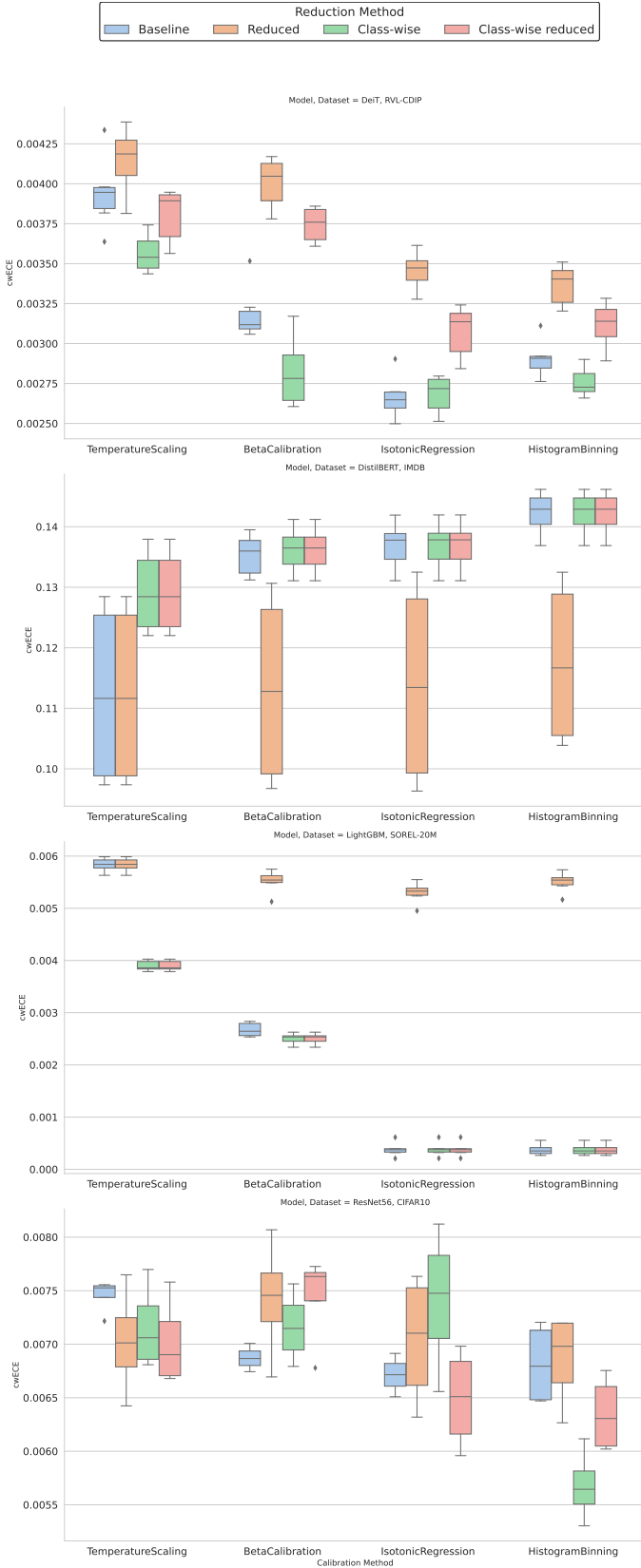

**Figure 6.** Comparison of beta calibration, isotonic regression, histogram binning and temperature scaling on cwECE for 4 models trained on 4 different real world datasets (6 folds; 25 bins).

| Model | Method | Reduced | | | Weighted Reduced | | |
|---|---|---|---|---|---|---|---|
| | | ECE | cwECE | $1-\delta$ | ECE | cwECE | $1-\delta$ |
| DeiT | Beta | 0.0192 | 0.0040 | 1.00 | 0.0176 | 0.0035 | 0.97 |
| | Histogram | 0.0049 | 0.0033 | 0.99 | 0.0052 | 0.0030 | 0.99 |
| | Isotonic | 0.0058 | 0.0034 | 0.99 | 0.0057 | 0.0031 | 0.99 |
| | TempScaling | 0.0255 | 0.0041 | 1.00 | 0.0251 | 0.0038 | 0.98 |
| ResNet56 | Beta | 0.0111 | 0.0074 | 1.00 | 0.0117 | 0.0076 | 0.97 |
| | Histogram | 0.0059 | 0.0068 | 0.99 | 0.0074 | 0.0069 | 0.98 |
| | Isotonic | 0.0109 | 0.0070 | 0.99 | 0.0110 | 0.0076 | 0.98 |
| | TempScaling | 0.0118 | 0.0070 | 1.00 | 0.0118 | 0.0075 | 1.00 |
| RF Balanced | Beta | 0.0110 | 0.0093 | 0.99 | 0.0110 | 0.0121 | 0.99 |
| | Histogram | 0.0069 | 0.0081 | 1.00 | 0.0069 | 0.0110 | 1.00 |
| | Isotonic | 0.0076 | 0.0086 | 1.00 | 0.0076 | 0.0117 | 1.00 |
| | TempScaling | 0.1362 | 0.0539 | 0.93 | 0.1046 | 0.0433 | 0.87 |
| RF Imbalanced | Beta | 0.0124 | 0.0127 | 0.99 | 0.0124 | 0.0162 | 0.99 |
| | Histogram | 0.0080 | 0.0120 | 1.00 | 0.0080 | 0.0157 | 1.00 |
| | Isotonic | 0.0090 | 0.0122 | 0.99 | 0.0090 | 0.0161 | 0.99 |
| | TempScaling | 0.1146 | 0.0122 | 0.93 | 0.0869 | 0.0398 | 0.87 |

**Table 3.** Evaluation of conditions for Reduced and Weighted Reduced methods listed in Corollary-4 and Corollary-8 respectively and comparisons of ECE and cwECE values.

# BIBLIOGRAPHY

[1] J. v. d. Hoeven et al, "GNU TeXmacs," www.texmacs.org, 1998.

[2] M. Panchenko, L. Sticher, and A. Benmerzoug, "Kyle: A toolkit for classifier calibration," github.com/appliedAI-Initiative/kyle, appliedAI Initiative GmbH, jun 2021.

[3] A. Benmerzoug, "Experiments for class-wise and reduced calibration methods," github.com/appliedAI-Initiative/classwise-calibration-experiments, appliedAI Initiative GmbH, aug 2022.

[4] C. Guo, G. Pleiss, Y. Sun, and K. Q. Weinberger, "On Calibration of Modern Neural Networks," in *Proceedings of the 34th International Conference on Machine Learning*, (Sydney, Australia), pp. 1321–1330, jul 2017.

[5] V. Papyan, X. Y. Han, and D. L. Donoho, "Prevalence of Neural Collapse during the terminal phase of deep learning training," *Proceedings of the National Academy of Sciences*, vol. 117, no. 40, pp. 24652–24663, oct 2020.

[6] J. Pratt, "Must subjective probabilities be realized as relative frequencies". 1962.

[7] A. Niculescu-Mizil and R. Caruana, "Predicting good probabilities with supervised learning," in *Proceedings of the 22nd International Conference on Machine Learning - ICML '05*, (Bonn, Germany), pp. 625–632, ACM Press, 2005.

[8] M. Kull, M. Perello Nieto, M. Kängsepp, T. Silva Filho, H. Song, and P. Flach, "Beyond temperature scaling: Obtaining well-calibrated multiclass probabilities with Dirichlet calibration," in *Advances in Neural Information Processing Systems 32* (H. Wallach, H. Larochelle, A. Beygelzimer, F. d\textquotesingle Alché-Buc, E. Fox, and R. Garnett, eds.), pp. 12316–12326, Curran Associates, Inc., 2019.

[9] A. Kumar, P. S. Liang, and T. Ma, "Verified Uncertainty Calibration," in *Advances in Neural Information Processing Systems 32* (H. Wallach, H. Larochelle, A. Beygelzimer, F. d\textquotesingle Alché-Buc, E. Fox, and R. Garnett, eds.), pp. 3792–3803, Curran Associates, Inc., 2019.

[10] M. Kull, T. M. S. Filho, and P. Flach, "Beyond sigmoids: How to obtain well-calibrated probabilities from binary classifiers with beta calibration," *Electronic Journal of Statistics*, vol. 11, no. 2, pp. 5052–5080, 2017.

[11] P. G. Fonseca and H. D. Lopes, "Calibration of Machine Learning Classifiers for Probability of Default Modelling," Technical Report, James Finance (CrowdProcess Inc.), oct 2017.

[12] J. Vaicenavicius, D. Widmann, C. Andersson, F. Lindsten, J. Roll, and T. Schön, "Evaluating model calibration in classification," in *The 22nd International Conference on Artificial Intelligence and Statistics*, pp. 3459–3467, PMLR, apr 2019.

[13] X. Y. Han, V. Papyan, and D. L. Donoho, "Neural Collapse Under MSE Loss: Proximity to and Dynamics on the Central Path," in *International Conference on Learning Representations (ICLR2022)*, (Virtual event), 2022.

[14] D. Soudry, E. Hoffer, M. S. Nacson, S. Gunasekar, and N. Srebro, "The Implicit Bias of Gradient Descent on Separable Data," *Journal of Machine Learning Research*, vol. 19, no. 70, pp. 1–57, dec 2018.

[15] K. He, X. Zhang, S. Ren, and J. Sun, "Deep Residual Learning for Image Recognition," Technical Report, Microsoft, dec 2015.

[16] A. Krizhevsky, "Learning Multiple Layers of Features from Tiny Images," Technical Report, Canadian Institute for Advanced Research, aug 2009.

[17] H. Touvron, M. Cord, M. Douze, F. Massa, A. Sablayrolles, and H. Jégou, "Training data-efficient image transformers & distillation through attention," in *Proceedings of the 38th International Conference on Machine Learning*, vol. 139, pp. 10347–10357, PMLR, jan 2021.

[18] D. Lewis, G. Agam, S. Argamon, O. Frieder, D. Grossman, and J. Heard, "Building a test collection for complex document information processing," in *Proceedings of the 29th Annual International ACM SIGIR Conference on Research and Development in Information Retrieval - SIGIR '06*, (Seattle, Washington, USA), p. 665, ACM Press, 2006.

[19] A. W. Harley, A. Ufkes, and K. G. Derpanis, "Evaluation of deep convolutional nets for document image classification and retrieval," in *2015 13th International Conference on Document Analysis and Recognition (ICDAR)*, (Tunis, Tunisia), pp. 991–995, IEEE, aug 2015.

[20] V. Sanh, L. Debut, J. Chaumond, and T. Wolf, "DistilBERT, a distilled version of BERT: smaller, faster, cheaper and lighter," feb 2020.

[21] A. L. Maas, R. E. Daly, P. T. Pham, D. Huang, A. Y. Ng, and C. Potts, "Learning Word Vectors for Sentiment Analysis," *Proceedings of the 49th Annual Meeting of the Association for Computational Linguistics: Human Language Technologies*, pp. 142–150, jun 2011.

[22] G. Ke, Q. Meng, T. Finley, T. Wang, W. Chen, W. Ma, Q. Ye, and T.-Y. Liu, "LightGBM: A Highly Efficient Gradient Boosting Decision Tree," in *Advances in Neural Information Processing Systems*, vol. 30, (Long Beach, CA, USA), p. 9, dec 2017.

[23] R. Harang and E. M. Rudd, "SOREL-20M: A Large Scale Benchmark Dataset for Malicious PE Detection," dec 2020.

[24] C. Gupta and A. Ramdas, "Top-label calibration and multiclass-to-binary reductions," in *International Conference on Learning Representations (ICLR 2022)*, 2022.